\documentclass[conference]{IEEEtran}
\IEEEoverridecommandlockouts
\usepackage{cite}
\usepackage{amsmath,amssymb,amsfonts}
\usepackage{algorithmic}
\usepackage{graphicx}
\usepackage{textcomp}
\usepackage{xcolor}

\usepackage{booktabs}
\usepackage[ruled,vlined]{algorithm2e}
\usepackage{multirow}
\renewcommand{\footnoterule}{%
  \kern -3pt
  \hrule width \columnwidth height 0.4pt
  \kern 2pt
}

\newcommand{\oursol}{\textsc{ABD-Net}}

\def\BibTeX{{\rm B\kern-.05em{\sc i\kern-.025em b}\kern-.08em
    T\kern-.1667em\lower.7ex\hbox{E}\kern-.125emX}}
\begin{document}

\title{
Articulated-Body Dynamics Network: \\Dynamics-Grounded Prior for Robot Learning
}

\author{Sangwoo Shin\textsuperscript{1*}, Kunzhao Ren\textsuperscript{1}, Xiaobin Xiong\textsuperscript{2}, and Josiah P. Hanna\textsuperscript{1}
\thanks{\textsuperscript{1}University of Wisconsin--Madison. \textsuperscript{2}Shanghai Innovation Institute (SII). *Corresponding author: \texttt{sshin237@cs.wisc.edu}}
}

\maketitle

\begin{abstract}
Recent work in reinforcement learning has shown that incorporating structural priors for articulated robots, such as link connectivity, into policy networks improves learning efficiency.
However, dynamics properties, despite their fundamental role in determining how forces and motion propagate through the body, remain largely underexplored as an inductive bias for policy learning.
To address this gap, we present the Articulated-Body Dynamics Network ($\oursol$), a novel graph neural network architecture grounded in the computational structure of forward dynamics.
Specifically, we adapt the inertia propagation mechanism from the Articulated Body Algorithm, systematically aggregating inertial quantities from child to parent links in a tree-structured manner, while replacing physical quantities with learnable parameters.
Embedding $\oursol$ into the policy actor enables dynamics-informed representations that capture how actions propagate through the body, leading to efficient and robust policy learning.
Through experiments with simulated humanoid, quadruped, and hopper robots, our approach demonstrates increased sample efficiency and generalization to dynamics shifts compared to transformer-based and GNN baselines.
We further validate the learned policy on real Unitree G1 and Go2 robots, state-of-the-art humanoid and quadruped platforms, generating dynamic, versatile and robust locomotion behaviors through sim-to-real transfer with real-time inference.
\end{abstract}



\section{Introduction}
Across machine learning domains, incorporating domain-specific architectural priors has been a key driver of success.
For example, convolutional neural networks exploit translation equivariance in images, while transformers' attention mechanisms capture long-range dependencies in sequential data.
In reinforcement learning (RL) for robots, prior work has explored incorporating the robot's spatial structure into policy networks. 
Graph neural networks (GNNs) were a natural choice~\cite{nervenet, smp}, and more recently, transformer-based architectures have emerged as a competitive alternative by combining attention mechanisms with structural information~\cite{bot, swat, meta_morph}. 
These approaches leverage the geometric structure of the robot to impose message-passing order in GNNs, or define masking schemes and biases in attention mechanisms.
%


Despite their success, existing methods primarily use spatial structure to specify which components of the robot should exchange information (e.g., via adjacency or attention masks), but how information is aggregated and transformed between connected components carries no physical semantics and must be learned solely from data.
However, articulated systems possess a richer structure than mere connectivity: inertial quantities propagate from child to parent links along the kinematic tree, determining how each link's mass affects the motion of the entire chain.
This propagation structure, which fundamentally determines how rigid bodies move, is precisely what forward dynamics algorithms compute~\cite{recursive_crba, efficient_dynamic_computation, alternative_crba, featherstone_aba_paper}. 
This raises a natural question:
\begin{center}
\textbf{\textit{
Can the computational structure of forward dynamics, embedded directly into the policy architecture, serve as an effective inductive bias for learning control policies?
}}
\end{center}

\begin{figure}[t]
\centering
\includegraphics[width=\columnwidth]{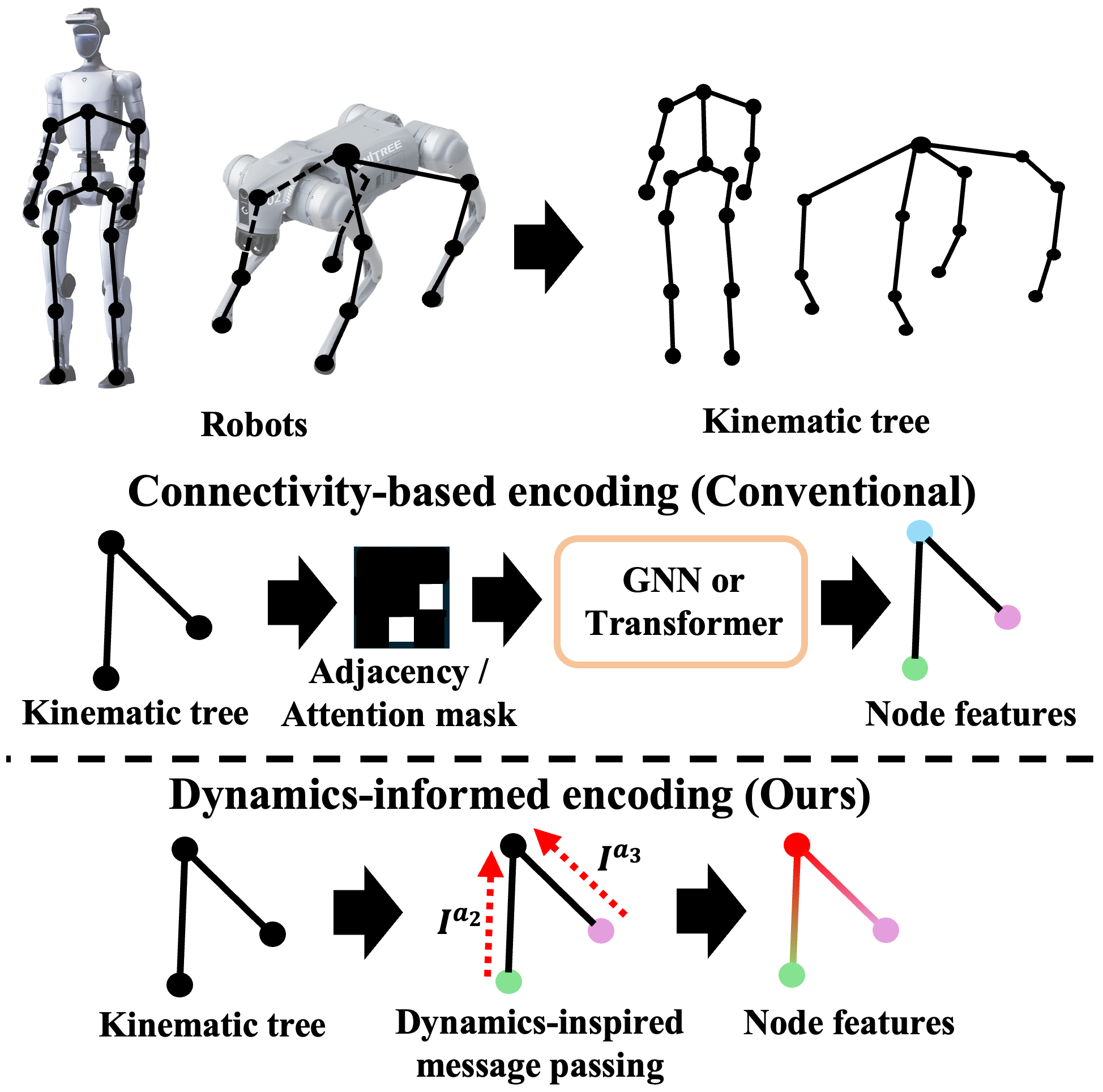}
\caption{
Different approaches to computing node features for articulated robots. Conventional methods use link connectivity to define information flow via adjacency or attention mask, leaving the network to learn how to form node features from scratch. 
Our approach encodes the computational structure of forward dynamics by performing dynamics-inspired message passing, where learnable inertia-related quantities (denoted as $I^a$) are propagated and aggregated from children to parents to form node features.
}
\label{fig: fig1}
\end{figure}

In this work, we investigate this question and answer it affirmatively by proposing \textbf{A}rticulated \textbf{B}ody \textbf{D}ynamics \textbf{Net}work $\oursol$, a graph neural network architecture that embeds the computational structure of forward dynamics into a policy actor network while replacing exact physical quantities with learnable representations (Fig.~\ref{fig: fig1}).
Specifically, $\oursol$ propagates features from child to parent links, mirroring how inertial effects accumulate in the Articulated Body Algorithm~\cite{featherstone_aba_paper}.
The resulting architecture not only provides an inductive bias rooted in dynamics, but also retains the flexibility to go beyond exact physical models. 
By structuring representations according to how physical quantities accumulate throughout the body, the actor can more directly infer how joint torques influence global motion, enabling effective and coordinated control actions.
This design distinguishes $\oursol$ from prior policy architectures~\cite{nervenet, smp, bot, swat}, which leverage the geometric connectivity of links but do not encode how features propagate along the kinematic tree.

While embedding forward dynamics into a neural network as a next-state predictor for model-based RL is natural, our objective differs: we investigate whether the forward-dynamics structure in the actor itself can improve policy learning efficiency, rather than building an accurate dynamics model.
We therefore consider model-free learning and embed the proposed structure directly into the actor network. This approach is compatible with the dominant paradigm in learning-based control, in which on-policy, model-free methods leverage massively parallel simulation.
We evaluate $\oursol$ on humanoid, quadruped, and hopper robots, demonstrating that our approach outperforms baselines in sample efficiency, generalization, and computational efficiency.

The main contributions of our work are as follows.
\textbf{\textit{(i)}} We propose \oursol, a novel graph neural network architecture that embeds the computational structure of forward dynamics into a policy actor.
\textbf{\textit{(ii)}} We provide reformulations of forward dynamics inertia propagation that avoid expensive matrix operations, enabling computationally efficient training.
\textbf{\textit{(iii)}} We empirically show that \oursol\ achieves superior sample efficiency, robustness under dynamics shifts, and computational efficiency across diverse morphologies and tasks. We further validate the learned policy through sim-to-real transfer on real robots, including Unitree G1 and Go2.

\section{Related Works}

\subsection{Exploiting Body Structure for Policy Learning}
GNNs have been adopted as policy architectures for robot control~\cite{pathak_gnn, nervenet, smp, gnn_for_physics_engine}. A common formulation represents each link and joint as graph nodes connected by edges that follow the robot's kinematic tree and performs learned message passing between physically adjacent components, leaving the semantics of feature aggregation to be inferred from data.
Building on this framework, NerveNet~\cite{nervenet} applies GNNs to design general RL actors, DGN~\cite{pathak_gnn} extends the idea to modular RL by allowing dynamically assembled actors, and SMP~\cite{smp} employs GNNs for morphology-agnostic policy learning.
More recently, transformer-based actors have emerged as a strong alternative. With variable context length, transformers naturally support fully connected graphs and have been shown to outperform GNNs even without explicit morphological priors~\cite{amorpheus, meta_morph}.
Subsequent works inject morphological knowledge through a traversal-order-based attention bias~\cite{swat} or connectivity-based attention masking~\cite{bot}.
More recently, MS-PPO~\cite{ms_ppo} encodes morphological symmetries, such as bilateral reflection, directly into a GNN policy to enforce equivariant behavior across symmetric gaits of quadrupeds.

While also leveraging body structure, $\oursol$ goes beyond connectivity by encoding the computational structure of forward dynamics, providing an explicit inductive bias for how features are transformed between connected links, applicable to arbitrary articulated morphologies.

\subsection{Physics-Informed Reinforcement Learning}
Physics-informed RL aims to incorporate physical structures or priors into the policy learning process~\cite{pirl_survey}. Such inductive biases can improve the sample efficiency of RL algorithms and enhance the safety of learned policies~\cite{pirl_safety_car, pirl_safety_invariance, pirl_safety_barrier}.
For example, Ramesh and Ravindran~\cite{pirl_mbrl} employ Lagrangian Neural Networks~\cite{lagrangian_nn} to learn the inertia tensor, which is then used with a numerical integrator to predict the next state, given analytically computed Coriolis and gravitational forces.
Rodrigues Network~\cite{rodrigues_net} is conceptually closest to our work in that it builds a neural operator from a classical robotics computation, by parameterizing the Rodrigues rotation formula within a transformer-style architecture.
While it demonstrates effectiveness in motion prediction and imitation learning, its extension to reinforcement learning remains unexplored.

$\oursol$ similarly incorporates physics-based inductive bias into the network architecture, but focuses on forward dynamics and applies this structure directly to policy learning in model-free RL, allowing the actor to implicitly learn how its actions propagate through the body. 

\begin{figure*}[t]
\centering
\includegraphics[width=\textwidth]{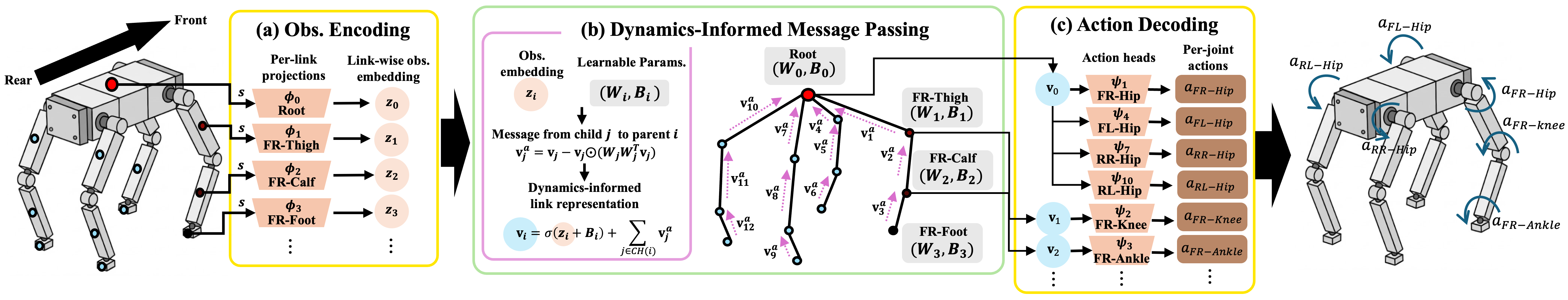}
\caption{
Overview of $\oursol$ on a quadruped robot. \textbf{(a) Observation Encoding:} Each link $i$ has its own projection $\phi_i$ that transforms the observation $\mathbf{s}$ into a link-wise observation embedding $\mathbf{z}_i$ (e.g., FR-Thigh, FR-Calf, FR-Foot for the front-right leg). \textbf{(b) Dynamics-Informed Message Passing:} Each link $i$ is associated with learnable parameters $(\mathbf{W}_i, \mathbf{B}_i)$. Link $j$ computes a contribution $\mathbf{v}^a_j$ using $\mathbf{W}_j$ and sends it to its parent, which aggregates contributions from all children to form its link representation $\mathbf{v}_i$ ($\sigma$ denotes softplus). \textbf{(c) Action Decoding:} For joint $j$ connecting $\textsc{pa}(j)$ and link $j$, the action $\mathbf{a}_j$ is computed from the parent's representation $\mathbf{v}_{\textsc{pa}(j)}$.
}
\label{fig:fig2}
\end{figure*}

\section{Preliminaries}
\subsection{Reinforcement Learning}
We consider Markov decision processes defined by $(\mathcal{S}, \mathcal{A}, p, r, d_0, \gamma)$, where $\mathcal{S}$ and $\mathcal{A}$ denote the state and action space, $p(\mathbf{s}' \mid \mathbf{s}, \mathbf{a})$ the transition density, $r(\mathbf{s}, \mathbf{a})$ the reward, $d_0$ the initial state distribution, and $\gamma \in [0, 1)$ the discount factor. The RL objective is to find a policy $\pi: \mathcal{S} \rightarrow \mathcal{A}$ that maximizes the expected sum of discounted rewards $\mathbb{E}_{\pi}\!\left[\sum_{t=0}^{\infty} \gamma^t r(\mathbf{s}_t, \mathbf{a}_t)\right]$.

\subsection{Robot Morphology and Graph Neural Networks}
We represent an articulated robot with $K$ links as a tree graph $\mathcal{G} = (\mathcal{V}, \mathcal{E})$. Each node $v_i \in \mathcal{V}$ for $i \in \{0, \ldots, K-1\}$ represents a link of the robot, and there is an edge $(v_i, v_j) \in \mathcal{E}$ if link $i$ and link $j$ are connected by a joint.
Designating node $v_0$ as the root (e.g., the torso or base link) induces a parent-child hierarchy: for each node $v_i$, let $\textsc{ch}(i)$ denote the set of its children, and for each non-root node $i$, let $\textsc{pa}(i)$ denote its unique parent.

In GNNs, each node $v_i$ is associated with a representation vector $\mathbf{v}_i$, which we refer to as the \textit{link representation} to reflect its association with a specific link in the robot’s morphology. GNNs update these representations through message aggregation:
\begin{align}
    \mathbf{m}_i &\leftarrow \sigma\left(\{\mathbf{v}_j : j \in \mathcal{N}_i\}\right), \quad \forall i \in \{0, \cdots, K-1\} \\
    \mathbf{v}_i &\leftarrow f_\theta(\mathbf{v}_i, \mathbf{m}_i), \quad \forall i \in \{ 0, \cdots, K-1\}
\end{align}
where $\mathcal{N}_i = \{\textsc{pa}(i)\} \cup \textsc{ch}(i)$ denotes the neighborhood of node $v_i$, $\sigma$ is an aggregation function (e.g., sum or mean), and $f_\theta$ is learned functions that define how neighboring nodes communicate and update their representations.

\subsection{Forward Dynamics}
The forward dynamics problem is to compute link accelerations given applied forces. The Articulated Body Algorithm (ABA)~\cite{featherstone_aba_paper} models the dynamics of each link as $f_i = I^A_i \alpha_i + b^A_i$, where $f_i$ is the spatial force acting on link $i$ and $\alpha_i$ is its spatial acceleration.
Here, $I^A_i$ is the articulated-body inertia, which represents the effective inertia of link $i$ accounting for all dynamic effects from its subtree. The term $b^A_i$ is the bias force, which captures velocity-dependent effects, such as Coriolis and centrifugal forces accumulated from the subtree.
Notably, once $I^A_i$ and $b^A_i$ are known, one can compute the acceleration of link $i$ without requiring explicit knowledge of each descendant link. In this sense, $I^A_i$ and $b^A_i$ serve as a compact representation that encapsulates the dynamic effects of the entire subtree rooted at link $i$.

ABA computes $I^A_i$ and $b^A_i$ recursively from leaves to root. Specifically, for each link $i$, the articulated-body inertia $I^A_i$ is obtained by combining its own rigid-body inertia $I_i$ with the contributions from its children:
\begin{equation}
\textstyle
    I^A_i = I_i + \sum_{j \in \textsc{ch}(i)} I^a_j.
    \label{eqn: aba}
\end{equation}
Here, $I^a_j$ is the contribution from child $j$:
\begin{equation}
    I^a_j = I^A_j - I^A_j S_j (S_j^\top I^A_j S_j)^{-1} S_j^\top I^A_j,
    \label{eqn: aba_contribution}
\end{equation}
where $S_j$ is the motion subspace matrix of the joint connecting link $j$ to its parent, whose columns span the directions of motion permitted by the joint. 
Eq.~\eqref{eqn: aba_contribution} subtracts inertia along the joint-permitted directions from the child's articulated-body inertia, so that only the constrained portion propagates to the parent.
The bias force $b^A_i$ is computed similarly.

\section{Our Approach}
%


\noindent\textbf{Architecture overview.}
In $\oursol$, the parameterized actor network $\pi_\theta: \mathcal{S} \rightarrow \mathcal{A}$ consists of three modules:
\begin{equation}
    \pi_\theta = \Psi \circ \mathcal{M} \circ \Phi,
\end{equation}
where $\Phi: \mathcal{S} \rightarrow \mathbb{R}^{K \times d}$ encodes the observation into link-wise embeddings $\{\mathbf{z}_i\}_{i=0}^{K-1}$, $\mathcal{M}: \mathbb{R}^{K \times d} \rightarrow \mathbb{R}^{K \times d}$ performs dynamics-informed message passing to produce link representations $\{\mathbf{v}_i\}_{i=0}^{K-1}$, and $\Psi: \mathbb{R}^{K \times d} \rightarrow \mathcal{A}$ decodes joint actions from link representations.
We assume each non-root link is connected to its parent via an actuated joint, and the action specifies the control signal (e.g., target position or torque) for each joint.
Fig.~\ref{fig:fig2} illustrates the overall architecture of $\oursol$.

\subsection{Link-wise Observation Encoding}
\label{subsec: perlink_obs_encoding}
Given observation $\mathbf{s}$, the observation encoder $\Phi$ transforms $\mathbf{s}$ into link-wise embeddings $\{\mathbf{z}_i\}_{i=0}^{K-1}$. $\Phi$ consists of per-link projections $\{\phi_i\}_{i=0}^{K-1}$, in which each $\mathbf{z}_i = \phi_i(\mathbf{s})$ is computed independently for link $i$.

\subsection{Dynamics-Informed Message Passing}
\label{subsec: dynamics-informed message passing}
Given observation embeddings $\{\mathbf{z}_i\}$, the message passing module $\mathcal{M}$ transforms them into link representations $\{\mathbf{v}_i\}$ by exploiting the computational structure of ABA.
To this end, we associate each link $i$ with two learnable parameters:
$\mathbf{B}_i \in \mathbb{R}^{d}$, a base feature analogous to the rigid-body inertia $I_i$ in Eq.~\eqref{eqn: aba};
$\mathbf{W}_i \in \mathbb{R}^{d \times d}$, a motion basis analogous to the motion subspace $S_j$ in Eq.~\eqref{eqn: aba_contribution}.

\noindent\textbf{Bottom-Up message passing.} $\mathcal{M}$ aggregates messages only from children $\textsc{ch}(i)$, mirroring the leaf-to-root computation of ABA in Eq.~\eqref{eqn: aba}. Specifically, the message $\mathbf{m}_i$ is defined as:
\begin{equation}
\textstyle
    \mathbf{m}_i = \sum_{j \in \textsc{ch}(i)} \mathbf{v}^a_j,
    \label{eqn: abdnet_message}
\end{equation}
where $\mathbf{v}^a_j$ is the contribution from child $j$, analogous to $I^a_j$ in Eq.~\eqref{eqn: aba_contribution}. The link representation is then computed as:
\begin{equation}
    \mathbf{v}_i = \text{softplus}(\mathbf{z}_i + \mathbf{B}_i) + \mathbf{m}_i.
    \label{eqn: abdnet_node_update}
\end{equation}
We apply softplus to ensure positivity, reflecting the positive-definiteness of rigid-body inertia.

\noindent\textbf{Child contribution.}
We translate Eq.~\eqref{eqn: aba_contribution} into a learnable form by replacing the articulated-body inertia $I^A_j$ with $\text{diag}(\mathbf{v}_j)$ and the motion subspace $S_j$ with a learnable motion basis $\mathbf{W}_j$.
To avoid the numerically unstable inverse, we assume 
$(\mathbf{W}^\top \text{diag}(\mathbf{v}) \mathbf{W})^{-1} \approx \mathbf{I}$, 
which holds when $\mathbf{W}$ has orthonormal columns weighted by $\mathbf{v}$.
Retaining the projection structure $\mathbf{W}_j \mathbf{W}_j^\top$, this yields:
\begin{equation}
\textstyle
\label{eqn: abdnet_contribution}
    \mathbf{v}^a_j = \mathbf{v}_j - \mathbf{v}_j \odot (\textbf{W}_j \textbf{W}_j^\top \mathbf{v}_j),
\end{equation}
where $\odot$ is element-wise multiplication.

Overall, our formulation in Eq.~\eqref{eqn: abdnet_message}--\eqref{eqn: abdnet_contribution} explicitly encodes the computational structure through which physical quantities propagate in ABA.
The term $\mathbf{W}_j \mathbf{W}_j^\top$ in Eq.~\eqref{eqn: abdnet_contribution} serves as a learned approximation of the constraint elimination mechanism in Eq.~\eqref{eqn: aba_contribution}, identifying feature directions to attenuate during child-to-parent propagation, analogous to how ABA removes inertia along joint-permitted directions.
To maintain this projection structure while allowing adaptability, we introduce an auxiliary loss.
%

\noindent\textbf{Orthogonality constraint.}
As described above, we approximate the inverse term $(\mathbf{W}^\top \text{diag}(\mathbf{v}) \mathbf{W})^{-1}$ as identity to avoid numerical instability.
To encourage this approximation to hold, we regularize $\mathbf{W}_i$ via an auxiliary loss:
\begin{equation}
\textstyle
\label{eq: ortho_loss}
    \mathcal{L}_{\text{orth}} = \frac{1}{K} \sum_{i=0}^{K-1} \| \mathbf{W}_i^\top \text{diag}(\mathbf{v}_i) \mathbf{W}_i - \mathbf{I} \|_F^2.
\end{equation}
As a soft constraint, this loss guides the network toward structural correspondence to ABA while allowing $\mathbf{W}_j \mathbf{W}_j^\top$ to deviate when doing so benefits policy learning, balancing structural fidelity with task-specific adaptation.
$\mathcal{L}_{\text{orth}}$ is added to the PPO objective during training.


\subsection{Action Decoding}
\label{subsec: action_decoding}
Given link representations $\{\mathbf{v}_i\}_{i=0}^{K-1}$, the action decoder $\Psi$ outputs the action for each joint. $\Psi$ consists of per-joint action heads $\{\psi_j\}_{j=1}^{K-1}$, one for each actuated joint.
We define joint $j$ as the joint connecting link $j$ to its parent $\textsc{pa}(j)$. 
Since each link representation $\mathbf{v}_i$ encapsulates the dynamic effects aggregated from its descendants, we use $\mathbf{v}_{\textsc{pa}(j)}$ to generate the action for joint $j$. Specifically, the action for joint $j$ 
is computed as:
\begin{equation}
\textstyle
    \mathbf{a}_j = \psi_j(\mathbf{v}_{\textsc{pa}(j)}) \in \mathbb{R}^{n_j},
\end{equation}
where $n_j$ is the degrees of freedom of joint $j$.
The complete forward pass of $\oursol$ is outlined in Algorithm~\ref{alg:abdnet}.

\begin{algorithm}[t]
\caption{Forward pass of $\oursol$}
\label{alg:abdnet}
\begin{algorithmic}[1]
\REQUIRE Observation $\mathbf{s}$
\ENSURE Action $\mathbf{a}$
\STATE \textbf{\textit{/* Link-wise Observation Encoding (Sec.~\ref{subsec: perlink_obs_encoding}) */}}
\STATE $\mathbf{z}_i \leftarrow \phi_i(\mathbf{s})$ for all $i \in \{0, \ldots, K-1\}$
\STATE \textbf{\textit{/* Dynamics-Informed Message Passing (Sec.~\ref{subsec: dynamics-informed message passing}) */}}
\STATE $\mathbf{m}_i \leftarrow \mathbf{0}$ for all $i \in \{0, \ldots, K-1\}$
\FOR{$i$ in leaf-to-root traversal}
    \STATE $\mathbf{v}_i \leftarrow \text{softplus}(\mathbf{z}_i + \mathbf{B}_i) + \mathbf{m}_i$
    \IF{$i \neq 0$}
        \STATE $\mathbf{v}^a_i = \mathbf{v}_i - \mathbf{v}_i \odot (\textbf{W}_i \textbf{W}_i^\top \mathbf{v}_i)$
        \STATE $\mathbf{m}_{\textsc{pa}(i)} \leftarrow \mathbf{m}_{\textsc{pa}(i)} + \mathbf{v}^a_i$
    \ENDIF
\ENDFOR
\STATE \textbf{\textit{/* Action Decoding (Sec.~\ref{subsec: action_decoding}) */}}
\RETURN $\mathbf{a} = \{\psi_i(\mathbf{v}_{\textsc{pa}(i)})\}_{i=1}^{K-1}$
\end{algorithmic}
\end{algorithm}

\section{Experiments}
We evaluate $\oursol$ in both simulation and real-world settings.
In simulation, we assess sample efficiency and generalization to dynamics shifts, and computational efficiency across diverse morphologies and tasks using two physics backends (Sec~\ref{subsec: simulation experiments}).
We then validate the learned policy on real Unitree G1 and Go2 robots,
demonstrating robust sim-to-real transfer with real-time onboard inference (Sec~\ref{subsec: sim2real}).

\begin{figure*}[t]
\vspace{-10pt}
\centering
\includegraphics[width=\textwidth]{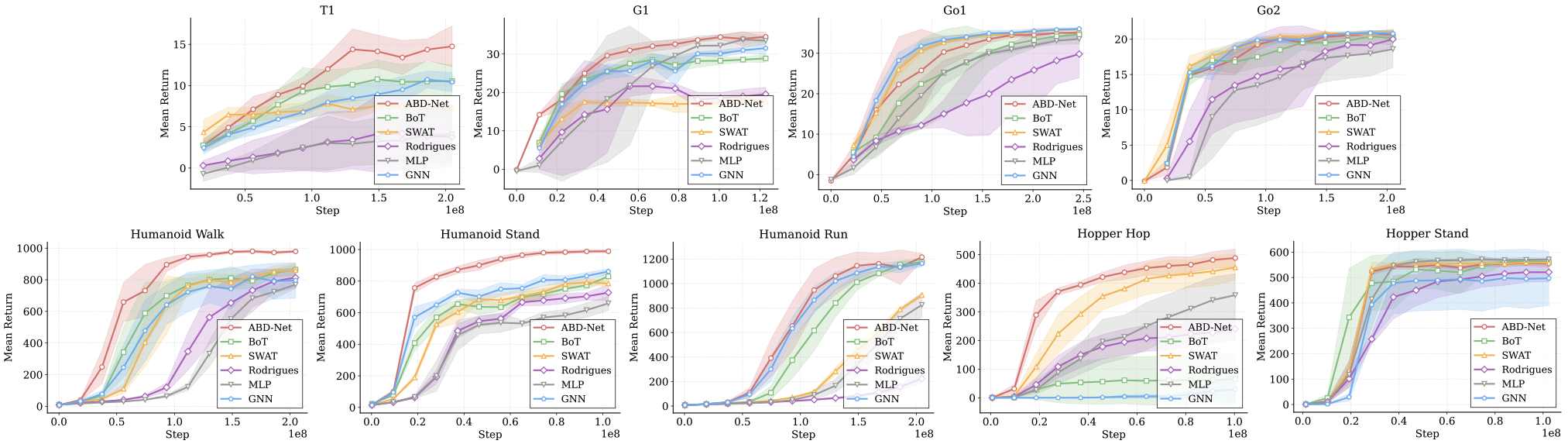}
\caption{Learning curves comparing ABD-Net (ours) and baselines: mean return versus the number of environment steps. All methods are trained with 5 seeds. Shaded regions indicate 95\% standard confidence intervals.}
\vspace{-5pt}
\label{fig: main_results}
\end{figure*}

\subsection{Simulation Experiments}
\label{subsec: simulation experiments}

\noindent \textbf{Environments.}
To evaluate $\oursol$ across different physics backends, we use two simulators: Genesis~\cite{Genesis} and SAPIEN~\cite{Sapien}.
In Genesis, we consider Booster T1, Unitree G1 (humanoid), Go1, and Go2 (quadruped) for velocity tracking tasks.
In SAPIEN, we use the MuJoCo Humanoid and Hopper from ManiSkill~\cite{maniskill3}. Humanoid tasks include Walk (moving along a target direction), Stand (rising from a random initial pose), and Run (moving forward at high speed). Hopper tasks include Hop (moving forward by hopping while staying upright) and Stand (remaining stationary and balanced).
\begin{table}[t]
\centering
\footnotesize
\setlength{\tabcolsep}{3pt}
\caption{Reward functions and weights for Genesis environments. ($\downarrow$) denotes penalty terms. }
\label{tab:reward_functions}
\renewcommand{\arraystretch}{0.85}
\begin{tabular}{llllll}
\toprule
Category & Reward Term & Go1 & Go2 & G1 & T1 \\
\midrule
\multirow{2}{*}{Vel. Track.}
& Lin. vel. tracking          & 1.5    & 1.0   & 1.0   & 1.0   \\
& Ang. vel. tracking           & 0.75   & 0.2   & 0.5   & 0.5   \\
\midrule
\multirow{3}{*}{Base Reg.}
& Base height                  & 12.5   & 50.0  & 10.0  & 20.0  \\
& Vertical lin. vel. ($\downarrow$)       & 2.0    & 1.0   & 2.0   & 1.0   \\
& Ang. vel. xy ($\downarrow$)             & 0.025  & --    & 0.05  & 0.025 \\
\midrule
\multirow{5}{*}{Joint Reg.}
& Action rate ($\downarrow$)              & 0.02   & 0.005 & 0.01  & 0.1   \\
& Default pose ($\downarrow$)             & 0.1    & 0.1   & --    & 0.1   \\
& Torques ($\downarrow$)                  & 1e-5   & 1e-5    & 1e-5  & 5e-5  \\
& Joint vel. ($\downarrow$)               & --     & --    & 1e-3  & 1e-4  \\
& Joint pos. limits ($\downarrow$)        & --     & --    & 5.0   & --    \\
\midrule
\multirow{2}{*}{Orientation}
& Orientation ($\downarrow$)              & 0.2    & 0.2    & 0.1   & 0.1   \\
& Hip deviation ($\downarrow$)            & 0.2    & --    & 1.0   & --    \\
\midrule
\multirow{7}{*}{Foot Reg.}
& Feet height                  & 0.05   & --    & --    & --    \\
& Feet air time                & 1.5    & --    & --    & --    \\
& Feet slip ($\downarrow$)                & --     & --    & 0.2   & 0.1   \\
& Feet roll ($\downarrow$)                & --     & --    & --    & 0.1   \\
& Feet yaw diff. ($\downarrow$)           & --     & --    & 1.0   & 1.0   \\
& Feet yaw mean dev. ($\downarrow$)       & --     & --    & 1.0   & 1.0   \\
& Feet distance ($\downarrow$)            & --     & --    & 1.0   & 1.0   \\
\midrule
\multirow{4}{*}{Other}
& Invalid contact ($\downarrow$)          & 1.0    & --    & --    & 1.0   \\
& Gait pattern                 & --     & --    & 0.18  & 6.0   \\
& Alive bonus                  & --     & --    & 0.15  & --    \\
& Base acc. ($\downarrow$)                & --     & --    & 1e-4  & 1e-4  \\
\bottomrule

\end{tabular}
\end{table}

\noindent\textbf{Implementation.}
Observations in Genesis tasks include proprioceptive information (joint positions and velocities, IMU angular velocities, projected gravities), velocity command signals, previous actions, and base linear velocities. For G1 and T1, we additionally include gait phase encoding and foot positions relative to the base. 
For SAPIEN tasks, we use the default observation spaces from ManiSkill:
Humanoid observations include joint positions and velocities, actuator forces,
and external forces, while Hopper observations consist of joint positions and
velocities.
For rewards, T1 and G1 use reward functions adapted from Booster Robotics and Unitree Gym, respectively. Go1 and Go2 use reward functions adapted from IsaacLab and Genesis, respectively. SAPIEN tasks use the default ManiSkill reward functions.
Detailed reward functions and weights for Genesis tasks are summarized in Table~\ref{tab:reward_functions}.

To implement $\oursol$, we represent each robot as a kinematic tree extracted
from its URDF or MJCF model file, parsing parent-child relationships between
links while excluding sensor-related components (e.g., camera, IMU) to focus
on the physical structure relevant to control.

\noindent \textbf{Baselines.}
To investigate how architectural inductive biases affect policy learning, we compare $\oursol$ with the following actor designs.
\textbf{\textsc{BoT}}~\cite{bot} is a transformer that uses link connectivity as an attention mask, alternating between masked and unmasked attention;
\textbf{\textsc{SWAT}}~\cite{swat} is a transformer that uses tree-traversal-based positional embeddings and graph-based attention biases;
\textbf{\textsc{Rodrigues}}~\cite{rodrigues_net} encodes the parent-to-child transformation in forward kinematics by extending the Rodrigues rotation formula into a learnable operator;
\textbf{\textsc{GNN}} is a standard graph neural network that aggregates messages from all neighbors including both parent and children;
\textbf{\textsc{MLP}} is a standard multi-layer perceptron.
All methods, including $\oursol$, are trained with PPO~\cite{ppo} using an identical MLP value network, with comparable parameters.

\begin{table}[t]
\vspace{-10pt}
\centering
\caption{Normalized final performance on Genesis and SAPIEN.}
\label{tab:main_results}
\begin{tabular}{lccc}
\toprule
\multicolumn{4}{c}{\textbf{Genesis (T1 + G1 + Go1 + Go2)}} \\
\midrule
Method & IQM & Median & Mean \\
\midrule
\textsc{ABD-Net} & \textbf{0.85} (0.73, 0.95) & \textbf{0.86} (0.70, 0.97) & \textbf{0.83} (0.75, 0.91) \\
\textsc{BoT} & 0.69 (0.56, 0.85) & 0.67 (0.54, 0.93) & 0.70 (0.58, 0.82) \\
\textsc{SWAT} & 0.79 (0.50, 0.94) & 0.83 (0.39, 0.95) & 0.75 (0.57, 0.89) \\
\textsc{Rodrigues} & 0.45 (0.17, 0.71) & 0.45 (0.04, 0.81) & 0.45 (0.25, 0.65) \\
\textsc{GNN} & 0.76 (0.65, 0.85) & 0.76 (0.65, 0.87) & 0.74 (0.67, 0.82) \\
\textsc{MLP} & 0.34 (0.07, 0.72) & 0.32 (0.05, 0.78) & 0.40 (0.19, 0.63) \\
\midrule
\multicolumn{4}{c}{\textbf{SAPIEN (Humanoid + Hopper)}} \\
\midrule
Method & IQM & Median & Mean \\
\midrule
\textsc{ABD-Net} & \textbf{0.97} (0.94, 0.98) & \textbf{0.97} (0.96, 0.98) & \textbf{0.94} (0.91, 0.97) \\
\textsc{BoT} & 0.66 (0.47, 0.80) & 0.67 (0.50, 0.88) & 0.59 (0.47, 0.70) \\
\textsc{SWAT} & 0.71 (0.62, 0.80) & 0.71 (0.67, 0.81) & 0.70 (0.64, 0.77) \\
\textsc{Rodrigues} & 0.44 (0.31, 0.51) & 0.45 (0.35, 0.54) & 0.41 (0.32, 0.49) \\
\textsc{GNN} & 0.73 (0.55, 0.87) & 0.76 (0.58, 0.92) & 0.64 (0.51, 0.76) \\
\textsc{MLP} & 0.56 (0.41, 0.70) & 0.57 (0.39, 0.68) & 0.55 (0.44, 0.65) \\
\bottomrule
\end{tabular}
\vspace{-5pt}
\end{table}

\begin{table}[t]
\vspace{-10pt}
\centering
\caption{Mass generalization performance (Retention, \%). \\
N/C indicates that the training policy did not converge.}
\label{tab:cross_domain}
\begin{tabular}{lcccc}
\toprule
Algorithm & Humanoid & Hopper & Go2 & T1 \\
\midrule
$\oursol$ & \textbf{91.1 $\pm$ 1.4} & \textbf{62.4 $\pm$ 0.5} & \textbf{82.4 $\pm$ 2.1} & \textbf{81.1 $\pm$ 3.5} \\
\textsc{BoT} & 17.0 $\pm$ 0.6 & N/C & 69.7 $\pm$ 2.3 & 57.0 $\pm$ 2.1 \\
\textsc{SWAT} & 88.9 $\pm$ 2.0 & 29.7 $\pm$ 1.0 & 81.4 $\pm$ 2.2 & 53.4 $\pm$ 1.6 \\
\textsc{Rodrigues} & 28.0 $\pm$ 1.7 & 9.1 $\pm$ 0.4 & 53.7 $\pm$ 3.1 & N/C \\
\textsc{GNN} & 54.9 $\pm$ 0.8 & N/C & 82.9 $\pm$ 2.2 & 69.9 $\pm$ 2.4 \\
\textsc{MLP} & 78.4 $\pm$ 2.9 & 55.4 $\pm$ 2.1 & 66.5 $\pm$ 2.7 & N/C \\
\bottomrule
\end{tabular}
\vspace{-5pt}
\end{table}


\noindent\textbf{Overall performance.}
Fig.~\ref{fig: main_results} compares the learning curves of $\oursol$ against baselines (\textsc{BoT}, \textsc{SWAT}, \textsc{Rodrigues}, \textsc{GNN}, \textsc{MLP}) in terms of average training return.
$\oursol$ consistently outperforms baselines in both sample efficiency and final performance, except on Go1/Go2 and Hopper Stand, where it achieves results comparable to other methods.
Importantly, the performance gap increases for more complex morphologies and tasks requiring dynamic behavior. In contrast, simpler morphologies with quasi-static tasks (Go1/Go2 velocity tracking, Hopper Stand) show smaller gains, suggesting that dynamics-aware inductive biases become more critical as task complexity increases.

Table~\ref{tab:main_results} summarizes the normalized final performance across both simulators. $\oursol$ achieves the highest IQM in both Genesis and SAPIEN, outperforming the strongest baseline \textsc{SWAT} by 7.6\% and 36.6\%, respectively.
We attribute this performance gain to the dynamics-informed message passing, which explicitly encodes how inertial effects propagate throughout the body, enabling more efficient learning of coordinated control.

\noindent\textbf{Mass generalization.}
Table~\ref{tab:cross_domain} evaluates the robustness of learned policies to dynamics shifts. 
Specifically, we deploy trained policies on modified morphologies with increased base mass: 1.5--2.0$\times$ for MuJoCo Humanoid, Go2, and Hopper, and 1.1--1.5$\times$ for T1. We report the retention rate (percentage of final performance) with 95\% confidence interval computed from 500 evaluation episodes.
$\oursol$ achieves the highest retention rate across all morphologies, with an average improvement of $23.9$\% over the strongest baseline, SWAT.

\begin{figure}[t]
\centering
\includegraphics[width=0.9\columnwidth]{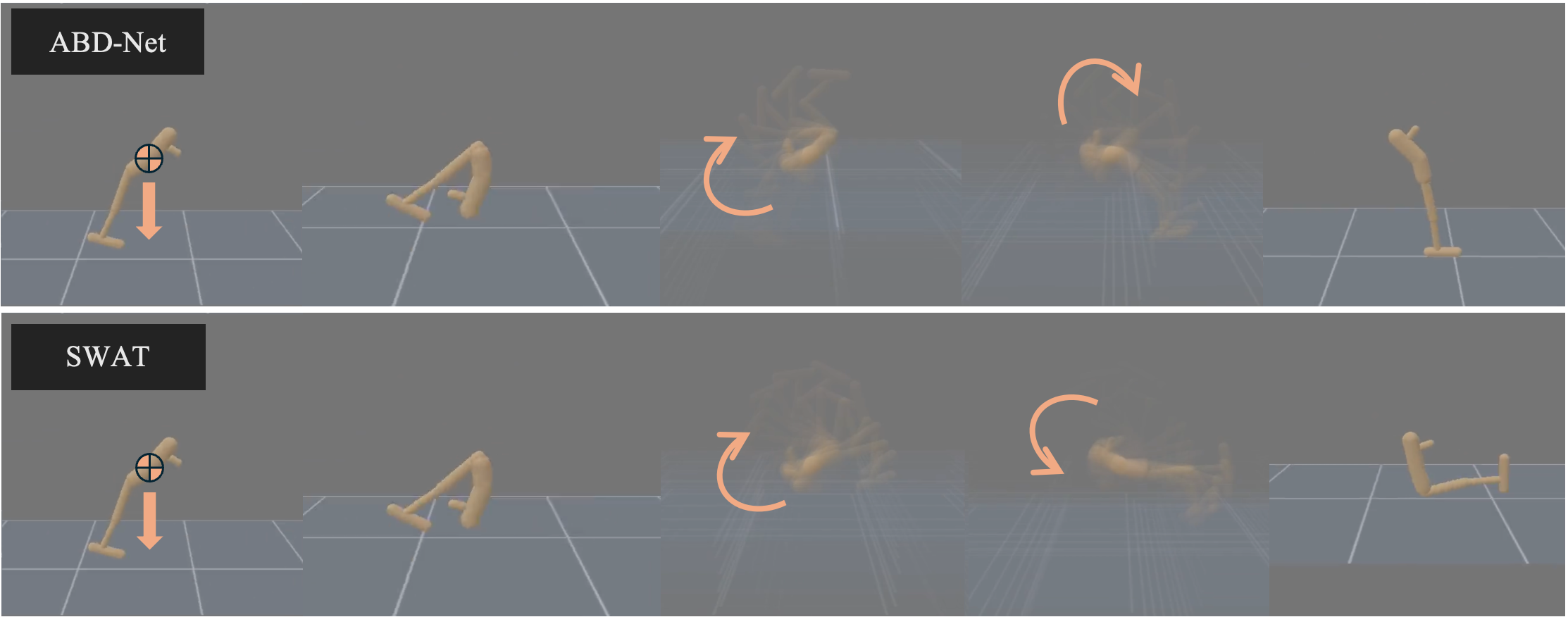}
\caption{Recovery behavior under 2$\times$ mass on Hopper Hop. Top: $\oursol$ applies stronger torque to recover from the downward tilt. Bottom: SWAT fails to compensate and falls.}
\vspace{-10pt}
\label{fig:mass_generalization}
\end{figure}

Fig.~\ref{fig:mass_generalization} illustrates this difference qualitatively on the Hopper Hop task with 2$\times$ base mass. When the increased mass causes the body to tilt downward, $\oursol$ successfully applies a stronger corrective torque to complete the forward rotation and recover balance, while SWAT fails to generate sufficient force and falls backward.
This robustness arises from the dynamics-informed architecture of $\oursol$, which explicitly constrains feature propagation to follow inertial accumulation in forward dynamics.
As a result, the policy leverages relative propagation structure rather than specific parameter values, yielding robustness to dynamics mismatch.

\begin{figure}[b]
\vspace{-10pt}
\centering
\includegraphics[width=1.0\columnwidth]{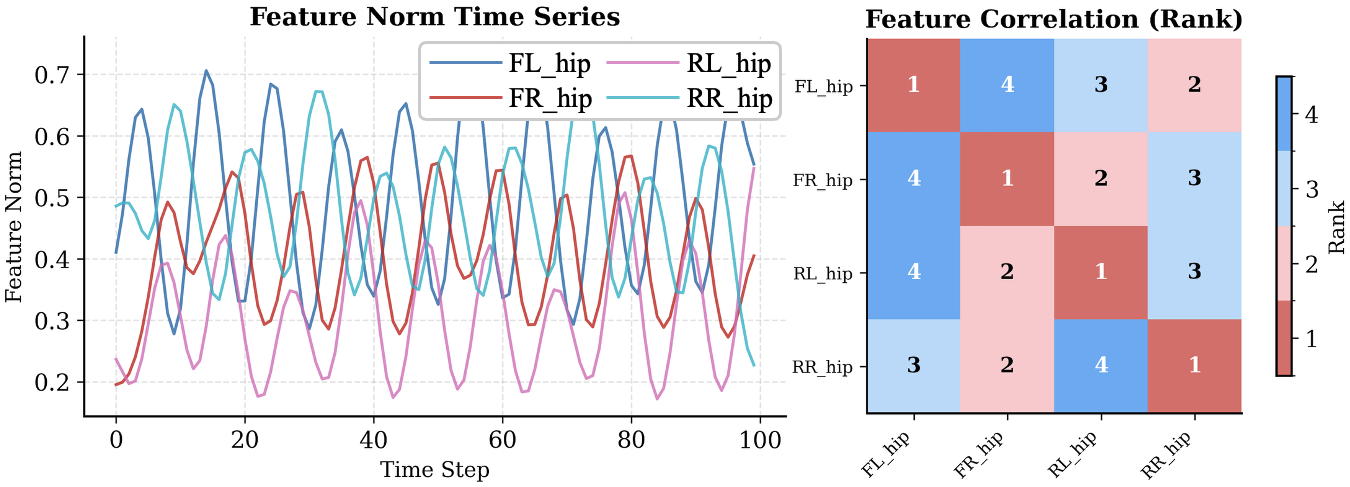}
\caption{Learned link representations on Go2 during trot gait. Left: Feature norm time series of hip joints (FL: front-left, RR: rear-right, etc.). Right: Correlation rank matrix between hip joint features, where lower rank indicates higher correlation.}
\label{fig:go2_feature_norm}
\vspace{-5pt}
\end{figure}

\noindent\textbf{Learned link representations.}
To examine whether the learned representations capture meaningful structure, we visualize the link representations of the hip joints (FL, FR, RL, RR) from a trained Go2 policy during locomotion in Fig.~\ref{fig:go2_feature_norm}. 

The learned policy exhibits a trot gait, where diagonal leg pairs (FL-RR, FR-RL) move in synchrony. The left plot shows that feature norms of diagonal pairs oscillate together over time. The right plot confirms this pattern through correlation analysis: each hip joint shows the highest correlation (excluding itself) with its diagonal counterpart (e.g., FL with RR, FR with RL), with RR\_hip being a minor exception.
This correspondence indicates that the dynamics-informed message passing in $\oursol$ learns link representations that capture physically meaningful relationships between body parts while remaining effective for task performance.

\begin{figure}[t]
\centering
\includegraphics[width=0.95\columnwidth]{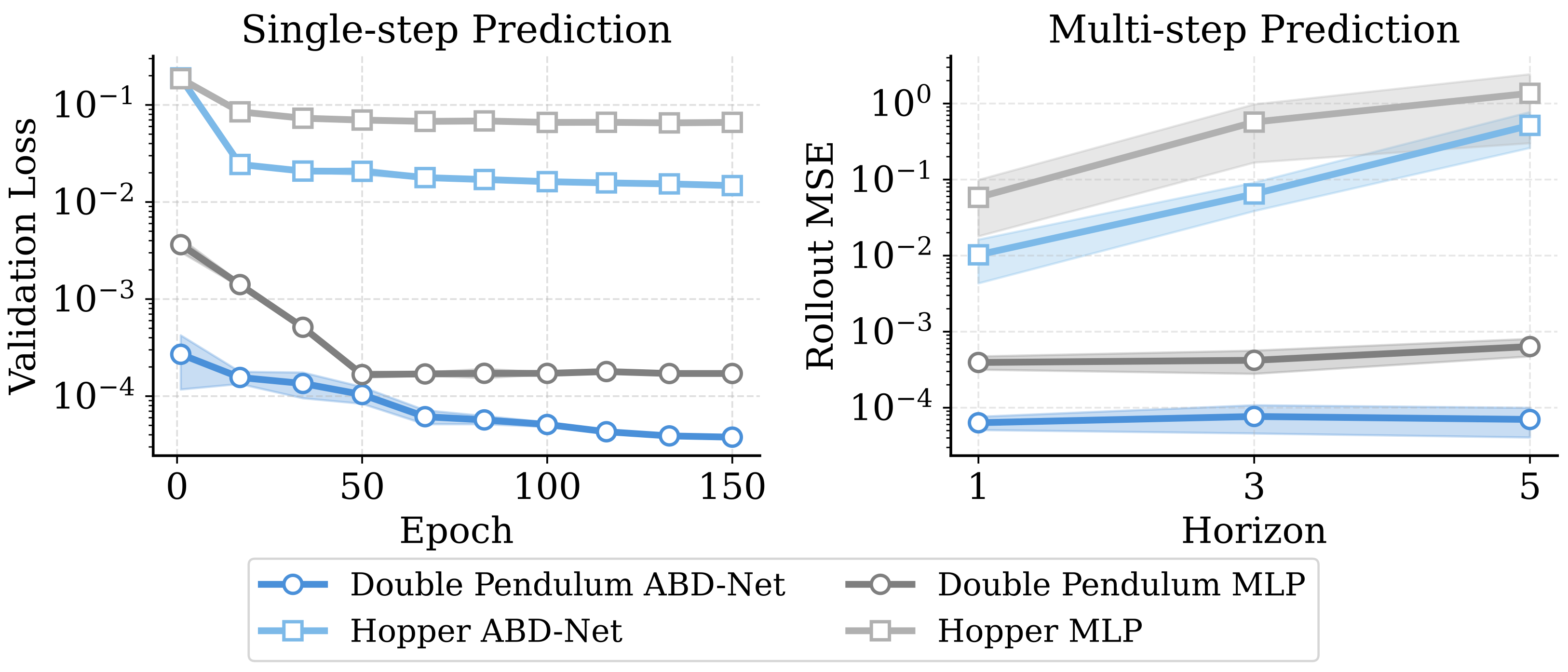}
\caption{Comparison of $\oursol$ and MLP as dynamics models. Left: Single-step validation loss on Double Pendulum and Hopper. Right: 1, 3, 5-step rollout prediction error.}
\label{fig:motivation}
\vspace{-10pt}
\end{figure}

\noindent\textbf{Extension to dynamics modeling.}
While $\oursol$ is designed for model-free policy learning, its architecture naturally extends to model-based settings.
To examine this, we train a dynamics model based on $\oursol$ and an MLP baseline to predict next observations given current observations and actions, i.e., $(\mathbf{s}, \mathbf{a}) \mapsto \mathbf{s}'$.
As shown in Fig.~\ref{fig:motivation}, the $\oursol$-based model achieves 10$\times$ lower validation loss than the MLP baseline in multi-step rollout prediction on both Double Pendulum (contact-free) and Hopper (contact-involved) environments.
This result demonstrates that the ABA-inspired architecture learns representations that capture physically meaningful state transitions, explaining its effectiveness in policy learning while also suggesting potential for model-based extensions such as model-based RL.

%
\noindent\textbf{Ablation studies.} We conduct ablation studies to isolate the effect of each architectural component.

\begin{figure}[b]
\vspace{-10pt}
\centering
\includegraphics[width=0.9\columnwidth]{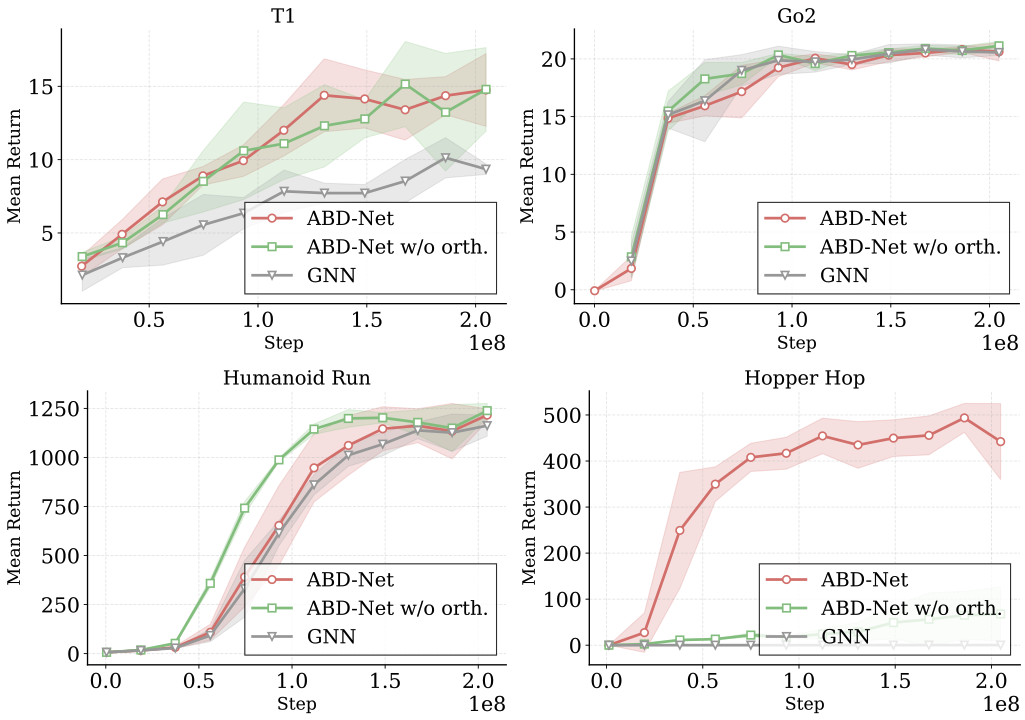}
\vspace{-5pt}
\caption{Learning curves comparing $\oursol$, $\oursol$ w/o orth., and GNN across four tasks}
\label{fig:ablation}
\vspace{-5pt}
\end{figure}

\noindent\textit{\textbf{Effect of dynamics-grounded constraints.}} To verify the contributions of the two key components in $\oursol$—the structured projection $\mathbf{W}\mathbf{W}^\top$ with orthogonality constraint and the bottom-up message passing—we consider two ablations:
(1) \textit{w/o orth.}, which replaces $\mathbf{W}_j\mathbf{W}_j^\top$ in Eq.~\eqref{eqn: abdnet_contribution} with an unconstrained matrix and removes $\mathcal{L}_{\text{orth}}$ in Eq.~\eqref{eq: ortho_loss};
(2) \textit{\textsc{GNN}}, a graph neural network.
Fig.~\ref{fig:ablation} compares learning curves, and Table~\ref{tab:ablation} reports mass generalization performance.

\begin{table}[b]
\centering
\caption{Mass generalization performance}
\label{tab:ablation}
\begin{tabular}{lcccc}
\toprule
Algorithm & Humanoid & Hopper & Go2 & T1 \\
\midrule
$\oursol$ & \textbf{91.1 $\pm$ 1.4} & \textbf{62.4 $\pm$ 0.5} & \textbf{96.2 $\pm$ 0.6} & \textbf{81.1 $\pm$ 3.5} \\
\textsc{w/o orth.} & 59.8 $\pm$ 0.4 & 56.3 $\pm$ 0.8 & 77.4 $\pm$ 2.2 & 51.6 $\pm$ 2.2 \\
\textsc{GNN} & 54.9 $\pm$ 0.8 & N/C & 82.9 $\pm$ 2.2 & 69.9 $\pm$ 2.4 \\
\bottomrule
\end{tabular}
\end{table}

\begin{figure*}[t]
\centering
\includegraphics[width=\textwidth]{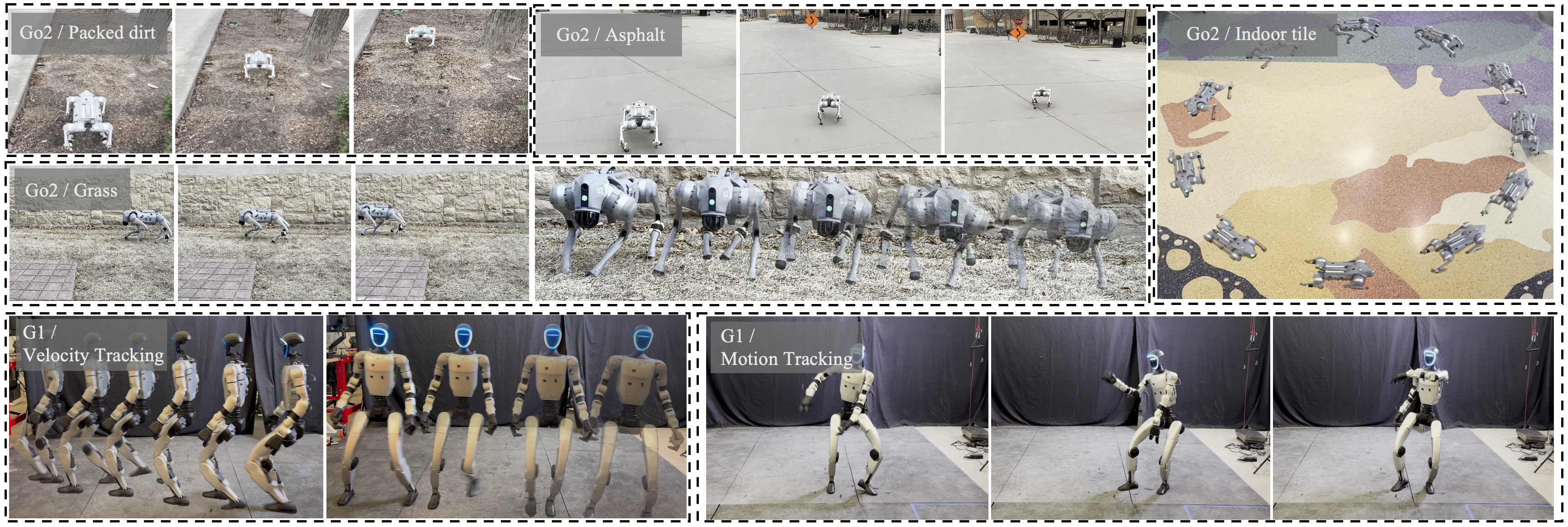}
\caption{Sim-to-real transfer on Unitree Go2 and G1 robots. Top two rows: Go2 velocity tracking across diverse terrains (packed dirt, asphalt, grass, indoor tile), with composite motion sequences showing lateral walking on grass and yaw tracking on indoor tile. Bottom: G1 velocity tracking with composite motion sequences showing forward and lateral walking (left) and motion tracking of a dance sequence (right).}
\label{fig:simtoreal}
\vspace{-5pt}
\end{figure*}

As shown, while \textit{w/o orth.} achieves comparable training performance to $\oursol$ on most tasks, the orthogonality constraint stabilizes learning in some cases, such as Hopper Hop.
Moreover, as shown in Table~\ref{tab:ablation}, the orthogonality constraint significantly improves generalization under dynamics shifts.
These results suggest that constraining policy structure to follow the propagation principles of forward dynamics leads to more reliable behavior under dynamics shifts.
\textit{\textsc{GNN}} fails to converge on T1 and Hopper Hop, indicating that bottom-up message passing aligned with ABA provides a stronger inductive bias than bidirectional propagation.
\color{black}

\begin{figure}[t]
\vspace{-10pt}
\centering
\includegraphics[width=0.58\columnwidth]{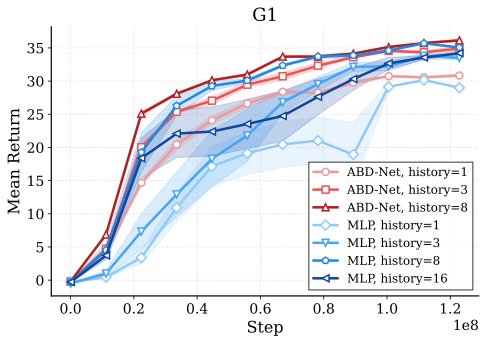}
\caption{Effect of observation history length on G1. $\oursol$ and \textsc{MLP} are compared across history lengths $\in \{1, 3, 8, 16\}$.}
\vspace{-12.5pt}
\label{fig:history_len}
\end{figure}

\noindent\textit{\textbf{Observation history length.}}
To assess whether the dynamics-informed architecture provides a more efficient prior than simply extending the observation history, we vary the history length $\in \{1, 3, 8, 16\}$ for both $\oursol$ and \textsc{MLP} on the G1 task (Fig.~\ref{fig:history_len}).
MLP improves with longer history ($1 \rightarrow 3 \rightarrow 8$), but degrades at $16$ due to the increased input dimensionality, indicating sensitivity to history length that may require task-specific tuning.
By contrast, $\oursol$ consistently outperforms \textsc{MLP} at each history length and exhibits smaller performance degradation at higher history lengths despite the increased input dimensionality.
These results suggest that the dynamics-informed architecture provides a structural prior that efficiently captures dynamics information with short histories, reducing the need for history length tuning.

\begin{figure}[t]
\vspace{-10pt}
\centering
\includegraphics[width=0.9\columnwidth]{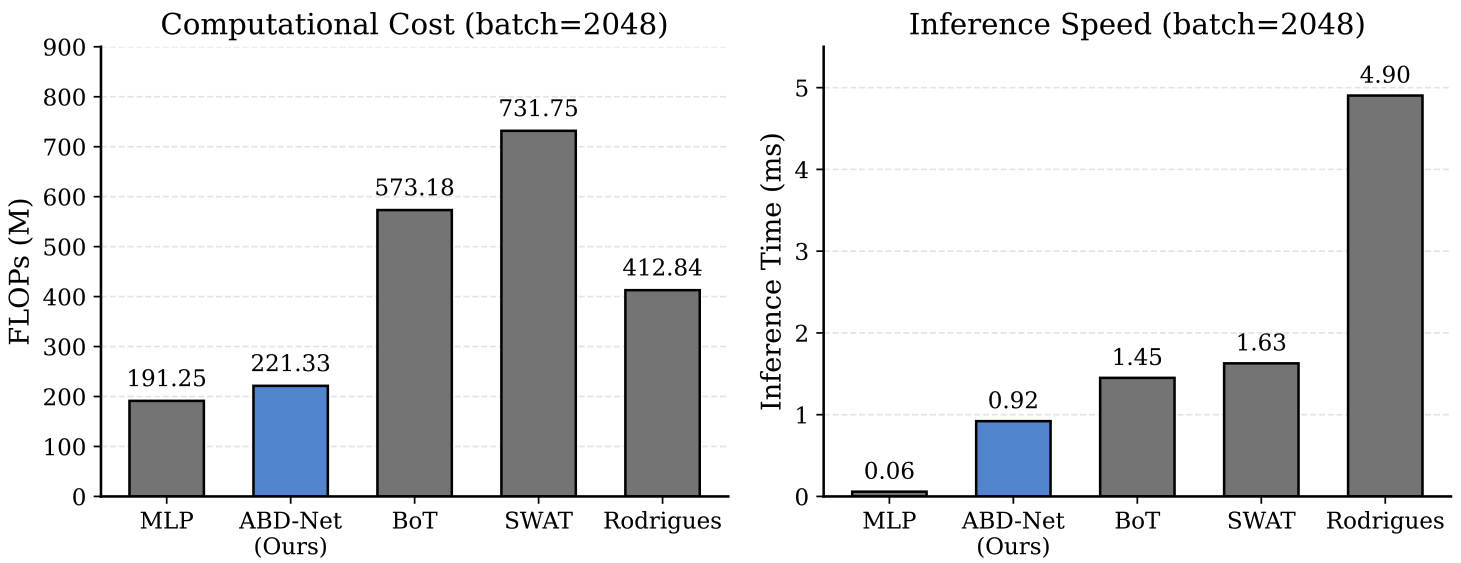}
\caption{Left: FLOPs per forward pass. Right: Inference time. All measured on an RTX 4090 with batch size 2048.}
\vspace{-12.5pt}
\label{fig:computation}
\end{figure}

\noindent\textbf{Computational efficiency.}
Fig.~\ref{fig:computation} compares computational cost across methods with similar parameter counts ($\sim$91--95K) on an NVIDIA RTX 4090.
To reflect the demands of large-scale parallel simulation training, we report forward throughput with batch size 2048.
$\oursol$ achieves 3$\times$ lower FLOPs than transformer-based methods while maintaining faster inference time.
Although $\oursol$ exhibits higher latency compared to the $\textsc{MLP}$ due to its sequential leaf-to-root computation, this overhead is marginal when considering the substantial gains in sample efficiency and robustness.
These results demonstrate that $\oursol$ strikes an effective balance between computational efficiency and task performance, making it suitable for modern parallelized RL frameworks.

\subsection{Hardware Experiments}
\label{subsec: sim2real}
To verify that $\oursol$ is suitable for real-world deployment, e.g., capable of real-time inference and producing stable behavior on hardware, we perform hardware experiments on Unitree G1 (humanoid) and Go2 (quadruped) robots.

\noindent\textbf{Training and deployment.}
We train velocity tracking and motion tracking policies using MJLab~\cite{mjlab}, an RL framework built on MuJoCo~\cite{mujoco}.
For G1 and Go2, we train velocity tracking policies on flat terrain; for G1, we additionally train a motion tracking policy to evaluate more dynamic behaviors, using a dance motion sequence provided by Unitree~\cite{unitree_rl_lab}. 
The simulation runs at 200\,Hz with a policy control frequency of 50\,Hz (decimation of 4).
The policy outputs desired joint positions, offset from the default standing pose. 
In hardware deployment, both policy inference and PD control run on the robot's onboard computer (NVIDIA Jetson Orin NX), with the policy running at 50\,Hz and the PD controller at 200\,Hz.
During deployment, velocity commands of up to $\pm$1.0\,m/s for forward and lateral directions and up to 0.5\,rad/s for yaw rate are used.
To bridge the sim-to-real gap, we apply domain randomization over foot friction, encoder bias, base center-of-mass offset, and external force perturbations.

\noindent\textbf{Results.}
As shown in Fig.~\ref{fig:simtoreal}, the Go2 achieves robust locomotion across diverse terrains including asphalt, packed dirt, grass, and indoor tile, and successfully follows forward, lateral, and yaw commands. The G1 also demonstrates forward and lateral walking as well as dynamic dancing behavior. 
These results confirm that \oursol{} is compatible with standard sim-to-real pipelines and capable of generating robust and dynamic motions with real-time onboard inference across different morphologies.
Furthermore, we verify that while $\oursol$ incurs higher inference latency than an MLP due to its sequential computation,
the worst-case onboard inference time remains under 5\,ms in the G1 motion tracking tasks,
well within the 20\,ms budget required for 50\,Hz control.
%

\section{Conclusion}
In this work, we presented $\oursol$, a graph neural network architecture that embeds the computational structure of forward dynamics into the policy actor for articulated robot control.
By propagating learnable features from child to parent links, mirroring how inertial quantities accumulate along the kinematic tree in the Articulated Body Algorithm, $\oursol$ provides a dynamics-grounded architectural inductive bias for policy learning.
Empirically, we demonstrated that $\oursol$ achieves increased sample efficiency, robustness under dynamics shifts, and computational efficiency across diverse morphologies and tasks, outperforming transformer-based and GNN baselines. We further validated the learned policy through sim-to-real transfer on real Unitree Go2 and G1 robots, confirming compatibility with standard deployment pipelines and real-time onboard inference.

\noindent\textbf{Limitations and future work.} 
While the worst-case inference latency remains well within real-time control budgets (Sec.~\ref{subsec: sim2real}), the sequential leaf-to-root computation in $\oursol$ incurs higher wall-clock training time compared to an MLP (4--6$\times$ in our PyTorch implementation), though recent results with a JAX implementation have reduced this gap to approximately 2$\times$.
Additionally, the current formulation operates on proprioceptive observations and does not directly handle high-dimensional sensory inputs such as images.

We plan to explore integration with model-based RL, where the dynamics-informed architecture can serve as both policy and dynamics model, potentially improving data efficiency further. We also aim to extend $\oursol$ to image-based observation settings and investigate its applicability to manipulation tasks involving contact-rich interactions.








\bibliographystyle{IEEEtran}
\bibliography{references}  

\vspace{12pt}



\end{document}